\documentclass[runningheads]{llncs}
\usepackage{graphicx}
\usepackage{comment}
\usepackage{amsmath,amssymb} 
\usepackage{color}
\usepackage{multirow}
\usepackage{color}
\usepackage{subfigure}
\usepackage{caption}
\usepackage{sidecap}

\sidecaptionvpos{figure}{c}

\begin{document}
\pagestyle{headings}
\mainmatter
\def\ECCVSubNumber{721}  

\title{Soft Anchor-Point Object Detection} 

\titlerunning{ECCV-20 submission ID \ECCVSubNumber} 
\authorrunning{ECCV-20 submission ID \ECCVSubNumber} 
\author{Anonymous ECCV submission}
\institute{Paper ID \ECCVSubNumber}

\titlerunning{Soft Anchor-Point Object Detection}
%
\author{Chenchen Zhu \and
Fangyi Chen \and
Zhiqiang Shen \and
Marios Savvides}
\authorrunning{C. Zhu et al.}
%
\institute{Carnegie Mellon University, Pittsburgh PA 15213, USA
\\
\email{\{chenchez, fangyic, zhiqians, marioss\}@andrew.cmu.edu}}
\maketitle

\begin{abstract}
Recently, anchor-free detection methods have been through great progress. The major two families, anchor-point detection and key-point detection, are at opposite edges of the speed-accuracy trade-off, with anchor-point detectors having the speed advantage. In this work, we boost the performance of the anchor-point detector over the key-point counterparts while maintaining the speed advantage. To achieve this, we formulate the detection problem from the anchor point's perspective and identify ineffective training as the main problem. Our key insight is that anchor points should be optimized jointly as a group both within and across feature pyramid levels. We propose a simple yet effective training strategy with soft-weighted anchor points and soft-selected pyramid levels to address the false attention issue within each pyramid level and the feature selection issue across all the pyramid levels, respectively. To evaluate the effectiveness, we train a single-stage anchor-free detector called Soft Anchor-Point Detector (SAPD). Experiments show that our concise SAPD pushes the envelope of speed/accuracy trade-off to a new level, outperforming recent state-of-the-art anchor-free and anchor-based detectors. Without bells and whistles, our best model can achieve a single-model single-scale AP of 47.4\% on COCO. 
\end{abstract}


\section{Introduction}

Recently, anchor-free object detectors \cite{cornernet,fsaf,guidedanchor,extremenet,fcos,foveabox,objectpoints,centernet,reppoints} have drawn a lot of attention. They don't rely on anchor boxes. Predictions are generated in a point(s)-to-box style. Compared to conventional anchor-based approaches \cite{faster-rcnn,ssd,yolov2,rfcn,dsod,dcn,fpn,retinanet,refinedet,cascade-rcnn,softer-nms}, anchor-free detectors have a few advantages in general: 1) no manual tuning of hyperparameters for the anchor configuration; 2) usually simpler architecture of detection head; 3) less training memory cost.

The anchor-free detectors can be roughly divided into two categories, i.e. anchor-point detection and key-point detection. Anchor-point detectors, such as \cite{densebox,unitbox,fsaf,guidedanchor,fcos,foveabox,reppoints}, encode and decode object bounding boxes as anchor points with corresponding point-to-boundary distances, where the anchor points are the pixels on the pyramidal feature maps and they are associated with the features at their locations just like the anchor boxes. Key-point detectors, such as \cite{cornernet,extremenet,centernet}, predict the locations of key points of the bounding box, e.g. corners, center, or extreme points, using a high-resolution feature map and repeated bottom-up top-down inference \cite{hourglass}, and group those key points to form a box. Compared to key-point detectors, anchor-point detectors have several advantages: 1) simpler network architecture; 2) faster training and inference speed; 3) potential to benefit from augmentations on feature pyramids \cite{m2det,libra-rcnn,hrnet}; 4) flexible feature level selection. However, they cannot be as accurate as key-point-based methods under the same image scale of testing.


A natural question to ask is: what hinders a simple anchor-point detector from achieving similar accuracy as key-point detectors?
In this work, we push the envelope further: we present Soft Anchor-Point Detector (SAPD), a concise single-stage anchor-point detector with both faster speed and higher accuracy. 
To achieve this, we formulate the detection problem from the anchor point's perspective and identify ineffective training as the major obstacle impeding anchor-point detector from exploring more potentials of network power both within and across the feature pyramid levels. Specifically, the conventional training strategy has two overlooked issues, i.e. false attention within each pyramid level and feature selection across all pyramid levels. For anchor points on the same pyramid level, those receiving false attention in training will generate detections with unnecessarily high confidence scores but poor localization during inference, suppressing some anchor points with accurate localization but lower score. This can confuse the post-processing step since high-score detections usually have priority to be kept over the low-score ones in non-maximum suppression, resulting in low AP scores at strict IoU thresholds. For anchor points at the same spatial location across different pyramid levels, their associated features are similar but how much they contribute to the network loss is decided without careful consideration. Current methods make the selection based on ad-hoc heuristics like instance scale and usually limited to a single level per instance. This causes a waste of unselected features.

These issues motivate us to propose a novel training strategy with two softened optimization techniques, i.e. soft-weighted anchor points and soft-selected pyramid levels. For anchor points on the same pyramid level, we reduce the false attention by reweighting their contributions to the network loss according to their geometrical relation with the instance box. We argue that the more close to the instance boundaries, the harder for anchor points to localize objects precisely due to feature misalignment, the less they should contribute to the network loss. Additionally, we further reweight an anchor point by the instance-dependent ``participation'' degree of its pyramid level. We implement a light-weight feature selection network to learn the per-level ``participation'' degrees given the object instances. The feature selection network is jointly optimized with the detector and not involved in detector inference.

\begin{SCfigure}
    \centering
    \includegraphics[width=0.6\columnwidth]{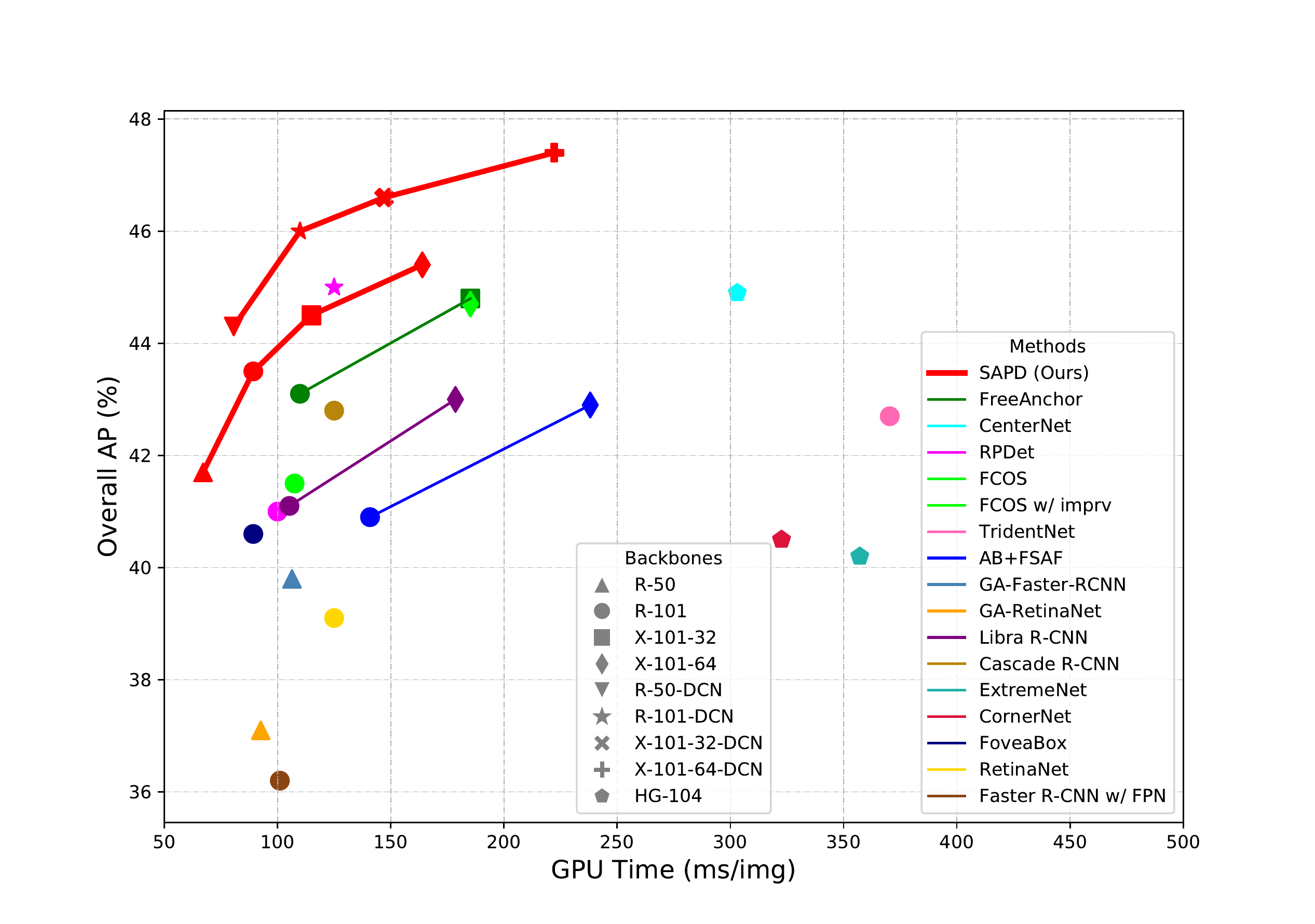}
    \caption{Single-model single-scale speed (ms) vs. accuracy (AP) on COCO \texttt{test-dev}. We show variants of our SAPD with and without DCN \cite{dcn}. Without DCN, our fastest version can run up to 5$\times$ faster than other methods with comparable accuracy. With DCN, our SAPD forms an upper envelop of all recent detectors.}
    \label{fig:ap-ms}
\end{SCfigure}

Comprehensive experiments show that the proposed training strategy consistently improves the baseline FSAF \cite{fsaf} module by a large margin \textit{without inference slowdown}, e.g. 2.1\% AP increase on COCO \cite{coco} detection benchmark with ResNet-50 \cite{resnet}. The improvements are robust and insensitive to specific hyperparameters and implementations, including advanced feature pyramid designs. With Balanced Feature Pyramid \cite{libra-rcnn}, our complete detector achieves the best speed-accuracy balance among recent state-of-the-art anchor-free detectors, see Figure \ref{fig:ap-ms}. We report single-model single-scale speed/accuracy of SAPD with different backbones, and with or without DCN \cite{dcn}. The fast variant without DCN outperforms the best key-point detector, CenterNet \cite{centernet} (45.4\% vs. 44.9\%) while running about 2$\times$ times faster. The accurate variant with DCN forms an upper bound of speed/accuracy trade-offs for recent single-stage and multi-stage detectors, surpassing the accurate TridentNet \cite{tridentnet} (47.4\% vs. 46.8\%) and being more than 3$\times$ faster.

\section{Related Work}
\label{sec:related}

\noindent \textbf{Anchor-free detectors}
Despite the dominance of anchor-based methods, anchor-free detectors are continuously under development. Earlier works like DenseBox \cite{densebox} and UnitBox \cite{unitbox} explore an alternate direction of region proposal. And it has been used in tasks such as scene text detection \cite{af-text} and pedestrian detection \cite{af-pedestrian}. Recent efforts have pushed the anchor-free detection outperforming the anchor-based counterparts. Most of them are single-stage detectors. For instance, CornerNet \cite{cornernet}, ExtremeNet \cite{extremenet} and CenterNet \cite{centernet} reformulate the detection problem as locating several key points of the bounding boxes. FSAF \cite{fsaf}, Guided Anchoring \cite{guidedanchor}, FCOS \cite{fcos} and FoveaBox \cite{foveabox} encode and decode the bounding boxes as anchor points and point-to-boundary distances. Anchor-free methods can also be in the form of two-stage detectors, such as Guided Anchoring \cite{guidedanchor} and RPDet \cite{reppoints}.

\noindent \textbf{Feature selection in detection}
Modern object detectors often construct the feature pyramid to alleviate the scale variation problem. With multiple levels in the feature pyramid, selecting the suitable feature level for each instance is a crucial problem. Anchor-based methods make the implicit selection by the anchor matching mechanism, which is based on ad-hoc heuristics like scales and aspect ratios. Similarly, most anchor-free approaches \cite{guidedanchor,fcos,foveabox,reppoints} assign the instances according to scale. The FSAF module \cite{fsaf}, on the other hand, makes the assignment by choosing the pyramid level with the minimal instance-dependent loss in a dynamic style during training but limited to one level per instance. In two-stage detectors, some methods consider feature selection in the second stage by feature fusion. PANet \cite{panet} proposed the adaptive feature pooling with the element-wise maximize operation. But this requires the input of region proposals for both training and testing, which is not compatible with single-stage methods. Our soft feature selection is designed for single-stage anchor-free methods and can dynamically choose multiple pyramid levels with differentiation.

\noindent \textbf{Soft weighting in detection}
FCOS \cite{fcos} predicts the ``center-ness'' masks and multiplies the confidence scores of anchor points with the masks. But this is in the inference stage so the false attention problem still affects the network training and the extra ``center-ness'' branches complicate the network architecture. We show that our simple soft-weighting scheme during training is more effective than the ``center-ness'' masks in the appendix. Previous works doing soft weighting in the training stage include Focal Loss \cite{retinanet} and Consistent Loss \cite{cl}. They reshape the classification loss but treat all samples independently. Our training strategy is more direct and comprehensive since we reshape the combination of classification and regression loss and consider jointly weighting a group of anchor points spreading both within and across feature pyramid levels.

\section{Soft Anchor-Point Detector}
\label{sec:sapd}
\begin{figure}
    \centering
    \includegraphics[width=\columnwidth]{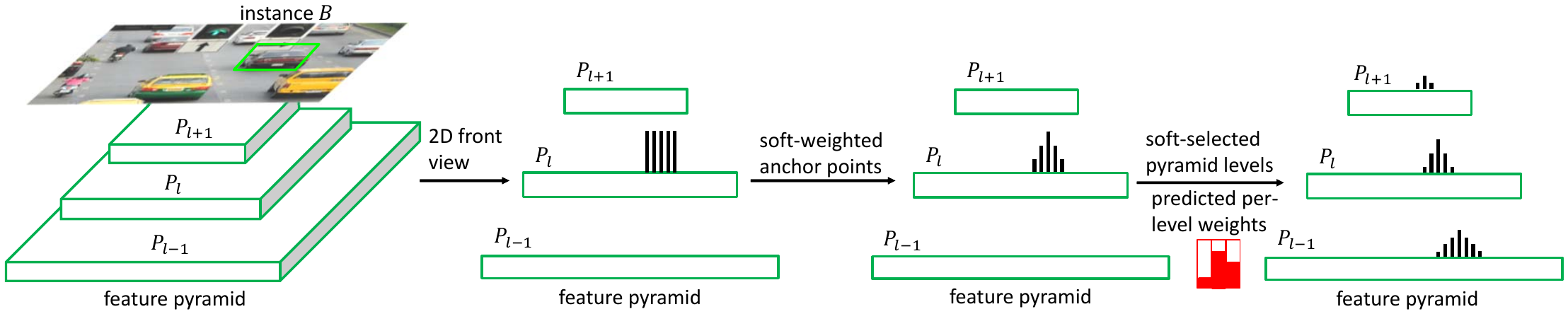}
    \caption{Illustrative overview of our training strategy with soft-weighted anchor points and soft-selected pyramid levels. The black bars indicate the assigned weights of positive anchor points' contribution to the network loss. The key insight is the joint optimization of anchor points as a group both within and across feature pyramid levels.}
    \label{fig:intro}
\end{figure}

In this section, we present our Soft Anchor-Point Detector (SAPD). First, we formulate the detection problem from the anchor point's perspective in the setting of a vanilla anchor-point detector with a simple head architecture (\ref{sec:sapd:preliminary}). Then we introduce our novel training strategy (Figure \ref{fig:intro}) including soft-weighted anchor points (\ref{sec:sapd:sw}) and soft-selected pyramid levels (\ref{sec:sapd:ss}) to address the false attention within pyramid level and feature selection across pyramid levels respectively.

\begin{figure}
    \centering
    \includegraphics[width=\columnwidth]{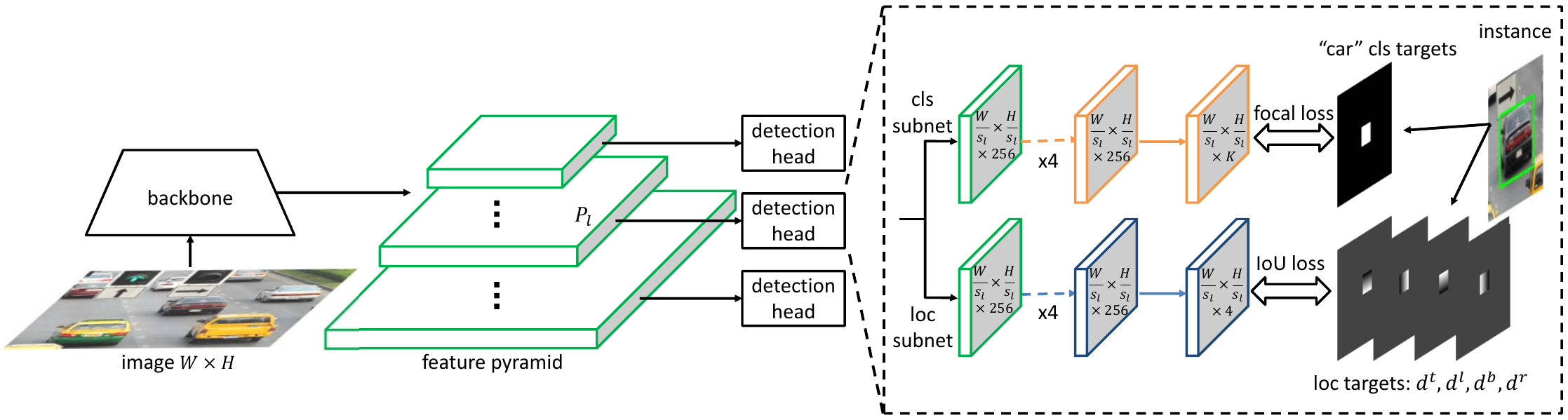}
    \caption{The network architecture of a vanilla anchor-point detector with a simple detection head.}
    \label{fig:preliminary}
\end{figure}

\subsection{Detection Formulation with Anchor Points}
\label{sec:sapd:preliminary}
The first anchor-point detector can be traced back to DenseBox \cite{densebox}. The recent modern anchor-point detectors are more or less attaching the detection head of DenseBox with additional convolution layers to multiple levels in the feature pyramids. Here we introduce the general concept of a representative in terms of network architecture, supervision targets, and loss functions.

\noindent \textbf{Network architecture} As shown in Figure \ref{fig:preliminary}, the network consists of a backbone, a feature pyramid, and one detection head per pyramid level, in a fully convolutional style. A pyramid level is denoted as $P_l$ where $l$ indicates the level number and it has $1 / s_l$ resolution of the input image size $W \times H$. $s_l$ is the feature stride and $s_l = 2^l$. A typical range of $l$ is 3 to 7. A detection head has two task-specific subnets, i.e. classification subnet and localization subnet. They both have five $3 \times 3$ conv layers. The classification subnet predicts the probability of objects at each anchor point location for each of the $K$ object classes. The localization subnet predicts the 4-dimensional class-agnostic distance from each anchor point to the boundaries of a nearby instance if the anchor point is positive (defined next).

\noindent \textbf{Supervision targets} We first define the concept of anchor points. An anchor point $p_{lij}$ is a pixel on the pyramid level $P_l$ located at $(i, j)$ with $i = 0, 1, \dots, W/s_l - 1$ and $j = 0, 1, \dots, H/s_l - 1$. Each $p_{lij}$ has a corresponding image space location $(X_{lij}, Y_{lij})$ where $X_{lij} = s_l(i + 0.5)$ and $Y_{lij} = s_l(j + 0.5)$. Next we define the valid box $B_v$ of a ground-truth instance box $B = (c, x, y, w, h)$ where $c$ is the class id, $(x, y)$ is the box center, and $w, h$ are box width and height respectively. $B_v$ is a central shrunk box of $B$, i.e. $B_v = (c, x, y, \epsilon w, \epsilon h)$, where $\epsilon$ is the shrunk factor. An anchor point $p_{lij}$ is positive if and only if some instance $B$ is assigned to $P_l$ and the image space location $(X_{lij}, Y_{lij})$ of $p_{lij}$ is inside $B_v$, otherwise it is a negative anchor point. For a positive anchor point, its classification target is $c$ and localization targets are calculated as the normalized distances $\mathbf{d} = (d^l, d^t, d^r, d^b)$ from the anchor point to the left, top, right, bottom boundaries of $B$ respectively \eqref{eq:loc_targets},
\begin{align}
\label{eq:loc_targets}
\begin{split}
    d^l &= \frac{1}{z s_l}[X_{lij} - (x - w/2)] \quad
    d^t = \frac{1}{z s_l}[Y_{lij} - (y - h/2)] \\ 
    d^r &= \frac{1}{z s_l}[(x + w/2) - X_{lij}] \quad
    d^b = \frac{1}{z s_l}[(y + h/2) - Y_{lij}] \\ 
\end{split}
\end{align}
where $z$ is the normalization scalar. For negative anchor points, their classification targets are background ($c=0$), and localization targets are set to null because they don't need to be learned. To this end, we have a classification target $c_{lij}$ and a localization target $\mathbf{d}_{lij}$ for all of each anchor point $p_{lij}$. A visualization of the classification targets and the localization targets of one feature level is illustrated in Figure \ref{fig:preliminary}.

\noindent \textbf{Loss functions}
Given the architecture and the definition of anchor points, the network generates a $K$-dimensional classification output $\mathbf{\hat{c}}_{lij}$ and a 4-dimensional localization output $\mathbf{\hat{d}}_{lij}$ per anchor point $p_{lij}$. Focal loss \cite{retinanet} ($l_{FL}$) is adopted for the training of classification subnets to overcome the extreme class imbalance between positive and negative anchor points. IoU loss \cite{unitbox} ($l_{IoU}$) is used for the training of localization subnets. Therefore, the per anchor point loss $L_{lij}$ is calculated as Eq. \eqref{eq:loss_ap}.
\begin{equation}
\label{eq:loss_ap}
    L_{lij} = 
    \begin{cases}
    l_{FL}(\mathbf{\hat{c}}_{lij}, c_{lij}) + l_{IoU}(\mathbf{\hat{d}}_{lij}, \mathbf{d}_{lij}), & p_{lij} \in p^+ \\
    l_{FL}(\mathbf{\hat{c}}_{lij}, c_{lij}), & p_{lij} \in p^-
    \end{cases}
\end{equation}
where $p^+$ and $p^-$ are the sets of positive and negative anchor points respectively. The loss for the whole network is the summation of all anchor point losses divided by the number of positive anchor points \eqref{eq:loss_net}.
\begin{equation}
\label{eq:loss_net}
    L = \frac{1}{N_{p^+}} \sum_l\sum_{ij} L_{lij}
\end{equation}

\subsection{Soft-Weighted Anchor Points}
\label{sec:sapd:sw}

\begin{figure}
    \begin{minipage}{.48\columnwidth}
    \centering
    \includegraphics[width=\columnwidth]{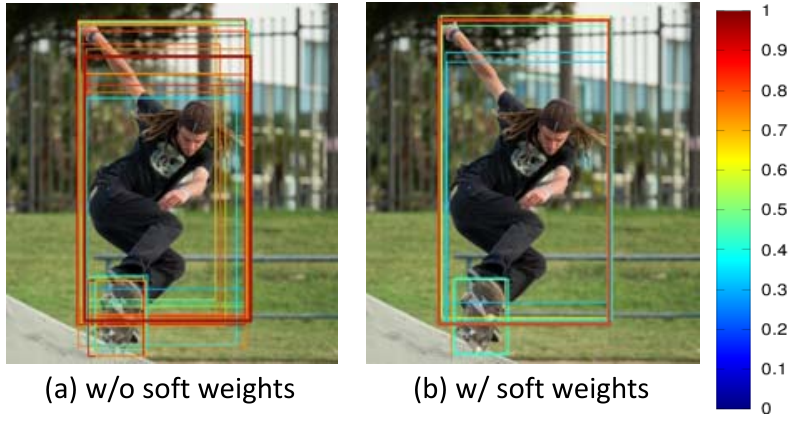}
    \caption{(a) Poorly localized detection boxes with high scores are generated by anchor points receiving false attention. (b) Our soft-weighting scheme effectively improves localization. Box score is indicated by the color bar.}
    \label{fig:softweight}
    \end{minipage}%
    \hfill
    \begin{minipage}{.48\columnwidth}
    \centering
    \includegraphics[width=\columnwidth]{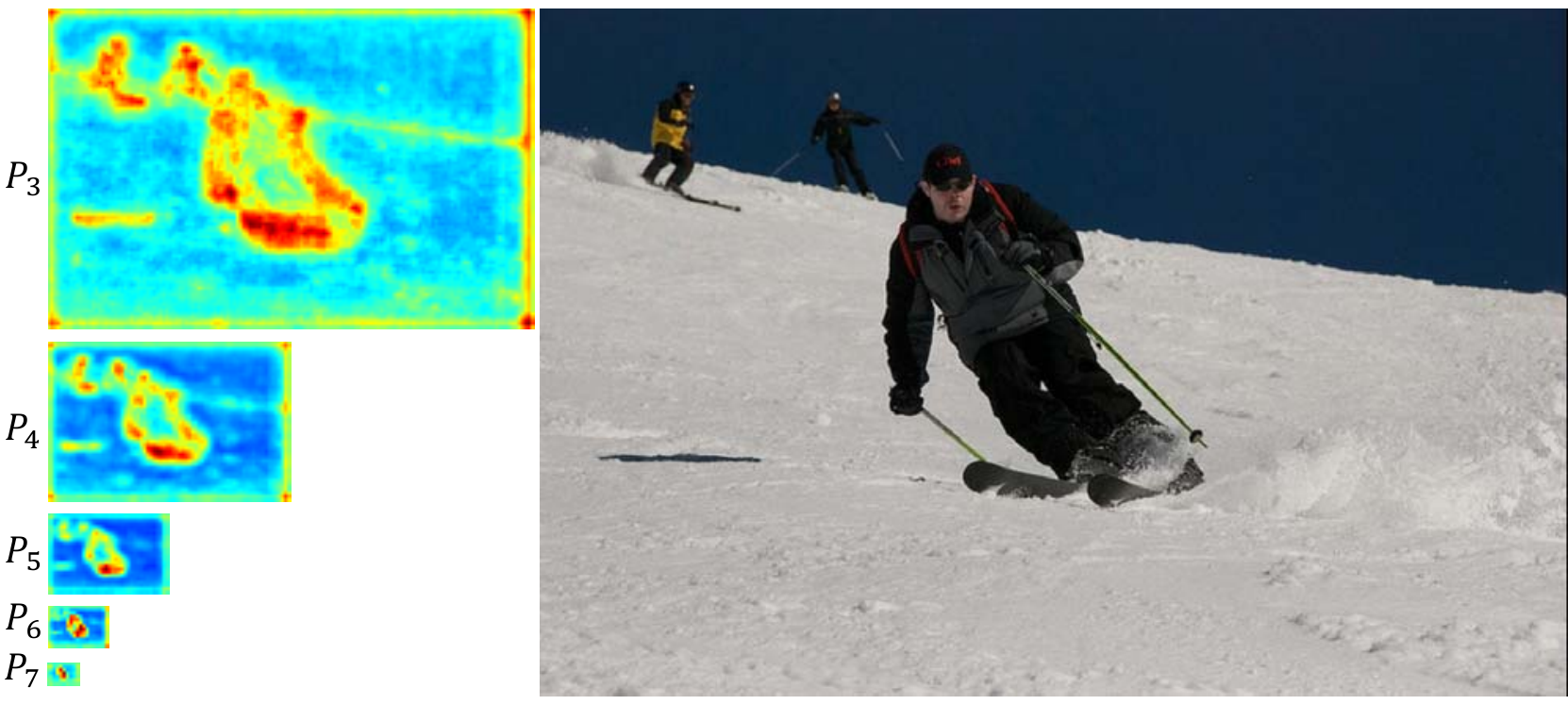}
    \caption{Feature responses from $P_3$ to $P_7$. They look similar but the details gradually vanish as the resolution becomes smaller. Selecting a single level per instance causes the waste of network power.}
    \label{fig:feat_response}
    \end{minipage}
\end{figure}

\noindent \textbf{False attention}  
Under the conventional training strategy, we observe that during inference some anchor points generate detection boxes with poor localization but high confidence score, which suppress the boxes with more precise localization but lower score. As a result, the non-maximum suppression (NMS) tends to keep the poorly localized detections, leading to low AP at a strict IoU threshold. 
We visualize an example of this observation in Figure \ref{fig:softweight} (a). We plot the detection boxes before NMS with confidence scores indicated by the color. The box with more precise localization of the person is suppressed by other boxes not so accurate but having high scores. Then the final detection (bold box) after NMS doesn't have high IoU with the ground-truth.

So why this is the case? The conventional training strategy treats anchor points independently in Eq. \eqref{eq:loss_net}, i.e. they receive equal attention. For a group of anchor points inside $B_v$, their spatial locations and associated features are different. So their abilities to localize $B$ are also different. We argue that anchor points located close to instance boundaries don't have features well aligned with the instance. Their features tend to be hurt by content outside the instance because their receptive fields include too much information from the background, resulting in less representation power for precise localization. Thus, forcing these anchor points to perform as well as those with powerful feature representation is misleading the network. Less attention should be paid to anchor points close to instance boundaries than those surrounding the center in training. In other words, the network should focus more on optimizing the anchor points with powerful feature representation and reduce the false attention to others.

\noindent \textbf{Our solution}
To address the false attention issue, we propose a simple and effective soft-weighting scheme. The basic idea is to assign an attention weight $w_{lij}$ for each anchor point's loss $L_{lij}$. For each positive anchor point, the weight depends on the distance between its image space location and the corresponding instance boundaries. The closer to a boundary, the more down-weighted the anchor point gets. Thus, anchor points close to boundaries are receiving less attention and the network focuses more on those surrounding the center. For negative anchor points, they are kept unchanged since they are not involved in localization, i.e. their weights are all set to 1. 
 Mathematically, $w_{lij}$ is defined in Eq. \eqref{eq:sw}:
\begin{equation}
\label{eq:sw}
    w_{lij} = 
    \begin{cases}
    f(p_{lij}, B), & \exists B, p_{lij} \in B_v \\
    1, & \text{otherwise}
    \end{cases}
\end{equation}
where $f$ is a function reflecting how close $p_{lij}$ is to the boundaries of $B$. Closer distance yields less attention weight. We instantiate $f$ using a generalized version of centerness function \cite{fcos}, i.e. $f(p_{lij}, B) = [\frac{\min(d_{lij}^l, d_{lij}^r) \min(d_{lij}^t, d_{lij}^b)}{\max(d_{lij}^l, d_{lij}^r) \max(d_{lij}^t, d_{lij}^b)}]^\eta$, where $\eta$ controls the decreasing steepness. An illustration of the soft-weighted anchor points is shown in Figure \ref{fig:intro}.


\subsection{Soft-Selected Pyramid Levels}
\label{sec:sapd:ss}
\noindent \textbf{Feature selection}
Unlike anchor-based detectors, anchor-free methods don't have constraints from anchor matching to select feature levels for instances from the feature pyramid. In other words, we can assign each instance to arbitrary feature level(s) in anchor-free methods during training. And selecting the right feature levels can make a big difference \cite{fsaf}. 

We approach the issue of feature selection by looking into the properties of the feature pyramid. Indeed, feature maps from different pyramid levels are somewhat similar to each other, especially the adjacent levels. We visualize the response of all pyramid levels in Figure \ref{fig:feat_response}. It turns out that if one level of feature is activated in a certain region, the same regions of adjacent levels may also be activated in a similar style. But the similarity fades as the levels are farther apart. This means that features from more than one pyramid level can participate together in the detection of a particular instance, but the degrees of participation from different levels should be somewhat different.

Inspired by the above analysis, we argue there should be two principles for proper pyramid level selection. Firstly, the selection should be related to the pattern of feature response, rather than some ad-hoc heuristics. And the instance-dependent loss can be a good reflection of whether a pyramid level is suitable for detecting some instances. This principle is also supported by \cite{fsaf}. Secondly, we should allow features from multiple levels involved in the training and testing for each instance, and each level should make distinct contributions. FoveaBox \cite{foveabox} has shown that assigning instances to multiple feature levels can improve the performance to some extent. But assigning to too many levels may instead hurt the performance severely. We believe this limitation is caused by the hard selection of pyramid levels. For each instance, the pyramid levels in FoveaBox are either selected or discarded. The selected levels are treated equally no matter how different their feature responses are.

Therefore, the solution lies in reweighting the pyramid levels for each instance. In other words, a weight is assigned to each pyramid level according to the feature response, making the selection soft. This can also be viewed as assigning a proportion of the instance to a level.


\begin{figure}
    \begin{minipage}{0.48\columnwidth}
    \centering
    \includegraphics[width=\columnwidth]{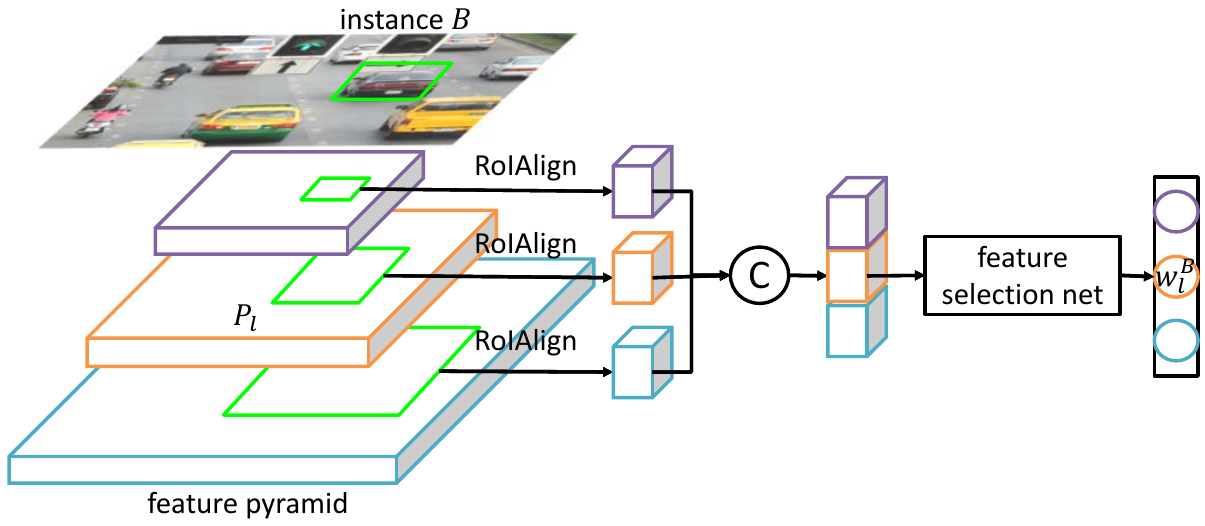}
    \caption{The weights prediction for soft-selected pyramid levels. ``C'' indicates the concatenation operation.}
    \label{fig:softselect}
    \end{minipage}
    \hfill
    \begin{minipage}{0.48\columnwidth}
    \scriptsize
    \centering
    \captionsetup{type=table}
    \begin{tabular}{c|c|c|c}
        \hline \hline
        layer type & output size & layer setting & activation \\ \hline
        input & $1280 \times 7 \times 7$ & n/a & n/a \\ 
        conv & $256 \times 5 \times 5$ & $3 \times 3, 256$ & relu \\
        conv & $256 \times 3 \times 3$ & $3 \times 3, 256$ & relu \\
        conv & $256 \times 1 \times 1$ & $3 \times 3, 256$ & relu \\
        fc & $5$ & n/a & softmax \\ \hline
    \end{tabular}
    \caption{Architecture of the feature selection network. The conv layers have no padding.}
    \label{tab:meta_net}
    \end{minipage}
\end{figure}

\noindent \textbf{Our solution}
So how to decide the weight of each pyramid level per instance? We propose to train a feature selection network to predict the weights for soft feature selection. The input to the network is instance-dependent feature responses extracted from all the pyramid levels. This is realized by applying the RoIAlign layer \cite{mask-rcnn} to each pyramid feature followed by concatenation, where the RoI is the instance ground-truth box. Then the extracted feature goes through a feature selection network to output a vector of the probability distribution, as shown in Figure \ref{fig:softselect}. We use the probabilities as the weights of soft feature selection.

There are multiple architecture designs for the feature selection network. For simplicity, we present a light-weight instantiation. It consists of three $3\times 3$ conv layers with no padding, each followed by the ReLU function, and a fully-connected layer with softmax. Table \ref{tab:meta_net} details the architecture. The feature selection network is jointly trained with the detector. Cross entropy loss is used for optimization and the ground-truth is a one-hot vector indicating which pyramid level has minimal loss as defined in the FSAF module \cite{fsaf}. 

So far, each instance $B$ is associated with a per level weight $w_l^B$ via the feature selection network. Together with the soft-weighting scheme in Section \ref{sec:sapd:sw}, the anchor point loss $L_{lij}$ is down-weighed further if $B$ is assigned to $P_l$ and $p_{lij}$ is inside $B_v$. We assign each instance $B$ to top$k$ feature levels with $k$ minimal instance-dependent losses during training. Thus, Eq. \eqref{eq:sw} is augmented into Eq. \eqref{eq:sw+ss}.
\begin{equation}
\label{eq:sw+ss}
    w_{lij} = 
    \begin{cases}
    w_l^B f(p_{lij}, B), & \exists B, p_{lij} \in B_v \\
    1, & \text{otherwise}
    \end{cases}
\end{equation}
The total loss of the whole model is the weighted sum of anchor point losses plus the classification loss ($L_{\text{select-net}}$) from the feature selection network, as in Eq. \eqref{eq:loss_total}. Figure \ref{fig:intro} shows the effect of applying soft-selection weights.
\begin{equation}
\label{eq:loss_total}
    L = \frac{1}{\sum_{p_{lij} \in p^+}w_{lij} } \sum_{lij} w_{lij}L_{lij} + \lambda L_{\text{select-net}}
\end{equation}
where $\lambda$ is the hyperparameter that controls the proportion of classification loss $L_{\text{select-net}}$ for feature selection.

\subsection{Implementation Details}
\noindent \textbf{Initialization}
We follow \cite{fsaf} for the initialization of the detection network. Specifically, the backbone networks are pre-trained on ImageNet1k \cite{imagenet}. The classification layers in the detection head are initialized with bias $-\log((1-\pi)/\pi)$ where $\pi = 0.01$, and a Gaussian weight. The localization layers in the detection head are initialized with bias 0.1, and also a Gaussian weight. For the newly added feature selection network, we initialize all layers in it using a Gaussian weight. All the Gaussian weights are filled with $\sigma = 0.01$.

\noindent \textbf{Optimization}
The entire detection network and the feature selection network are jointly trained with stochastic gradient descent on 8 GPUs with 2 images per GPU using the COCO \texttt{train2017} set \cite{coco}. Unless otherwise noted, all models are trained for 12 epochs ($\sim$90k iterations) with an initial learning rate of 0.01, which is divided by 10 at the 9th and the 11th epoch. Horizontal image flipping is the only data augmentation unless otherwise specified. 
For the first 6 epochs, we don't use the output from the feature selection network. The detection network is trained with the same online feature selection strategy as in the FSAF module \cite{fsaf}, i.e. each instance is assigned to only one feature level yielding the minimal loss. We plug in the soft selection weights and choose the top$k$ levels for the second 6 epochs. This is to stabilize the feature selection network first and to make the learning smoother in practice. We use the same training hyperparameters for the shrunk factor $\epsilon = 0.2$ and the normalization scalar $z = 4.0$ as \cite{fsaf}. We set $\lambda = 0.1$ although results are robust to the exact value.

\noindent \textbf{Inference}
At the time of inference, the network architecture is as simple as in Figure \ref{fig:preliminary}. The feature selection network is \textit{not} involved in the inference so the runtime speed is not affected. An image is forwarded through the network in a fully convolutional style. Then classification prediction $\mathbf{\hat{c}}_{lij}$ and localization prediction $\mathbf{\hat{d}}_{lij}$ are generated for all each anchor point $p_{lij}$. Bounding boxes can be decoded using the reverse of Eq. \eqref{eq:loc_targets}. We only decode box predictions from at most 1k top-scoring anchor points in each pyramid level, after thresholding the confidence scores by 0.05. These top predictions from all feature levels are merged, followed by non-maximum suppression with a threshold of 0.5, yielding the final detections.

\section{Experiments}
\label{sec:exp}
We conduct experiments on the COCO \cite{coco} detection track using the MMDetection \cite{mmdet} codebase. All models are trained on the \texttt{train2017} split including around 115k images. We analyze our method by ablation studies on the \texttt{val2017} split containing 5k images. When comparing to the state-of-the-art detectors, we report the Average Precision (AP) scores on the \texttt{test-dev} split. 

\begin{table}[]
\centering
\setlength\tabcolsep{5pt}
\begin{tabular}{c|c c c|c c c c c c}
\hline\hline
 & SW & SS & BFP & AP & AP$_{50}$ & AP$_{75}$ & AP$_{S}$ & AP$_{M}$ & AP$_{L}$ \\ \hline
FSAF \cite{fsaf} &  &  & & 35.9 & 55.0 & 37.9 & 19.8 & 39.6 & 48.2 \\ 
 & \checkmark &  &  & 37.0 & 55.8 & 39.5 & 20.5 & 40.1 & 48.5 \\ 
 & \checkmark & \checkmark &  & 38.0 & 56.9 & 40.5 & 21.2 & 41.2 & 50.2 \\ 
 & & & \checkmark & 36.8 & 57.4 & 38.8 & 21.6 & 40.9 & 47.6 \\
SAPD & \checkmark & \checkmark & \checkmark & \textbf{38.8} & \textbf{58.7} & \textbf{41.3} & \textbf{22.5} & \textbf{42.6} & \textbf{50.8} \\ \hline
\end{tabular}
\caption{Ablative experiments for the SAPD on the COCO \texttt{val2017}. ResNet-50 is the backbone network for all experiments in this table. We study the effect of \textbf{SW}: soft-weighted anchor points, \textbf{SS}: soft-selected pyramid levels, and \textbf{BFP} \cite{libra-rcnn}: augmented feature pyramids.}
\label{tab:ablation}
\end{table}

\subsection{Ablation Studies}
\label{sec:exp:ablation}
All results in ablation studies are based on models trained and tested with an image scale of 800 pixels. We study the contribution of each proposed component by gradually applying these components to the baseline FSAF \cite{fsaf} module. For the soft-weighted anchor points and soft-selected pyramid levels, we first study the effect of varying hyperparameters on them and then apply each component with the best hyperparameter to the baseline. We also report more ablative experiments in Appendix \ref{sec:app}.

\begin{table}[]
    \begin{minipage}{0.48\columnwidth}
    \centering
    \begin{tabular}{c|c c c c c c}
        \hline \hline
        $\eta$ & AP & AP$_{50}$ & AP$_{75}$ & AP$_S$ & AP$_M$ & AP$_L$ \\ \hline
        0.10 & 36.8 & 56.2 & 39.0 & 20.6 & 40.3 & 48.3\\
        0.50 & 36.9 & 55.8 & 39.4 & 20.2 & 40.1 & 48.7 \\
        1.0 & \textbf{37.0} & 55.8 & 39.5 & 20.5 & 40.1 & 48.5 \\
        2.0 & 36.6 & 55.1 & 39.1 & 19.8 & 40.3 & 47.8 \\
        \hline
    \end{tabular}
    \caption{Varying $\eta$ for the generalized centerness function in soft-weighted anchor points.}
    \label{tab:eta}
    \end{minipage}
    \hfill
    \begin{minipage}{0.48\columnwidth}
    \centering
    \begin{tabular}{c|c c c c c c}
        \hline \hline
        top$k$ & AP & AP$_{50}$ & AP$_{75}$ & AP$_S$ & AP$_M$ & AP$_L$ \\ \hline
        2 & 37.9 & 56.9 & 40.5 & 21.1 & 41.0 & 50.1 \\
        3 & \textbf{38.0} & 56.9 & 40.5 & 21.2 & 41.2 & 50.2 \\
        4 & 37.9 & 56.9 & 40.3 & 21.2 & 41.1 & 50.2 \\
        5 & 37.9 & 56.8 & 40.5 & 21.0 & 41.1 & 50.2 \\
        \hline
    \end{tabular}
    \caption{Varying $k$ for number of selected levels in soft-selected pyramid level with $\eta=1.0$.}
    \label{tab:topk}
    \end{minipage}
\end{table}

\textbf{Soft-weighted anchor points improve the localization.}
We first apply the soft-weighting scheme (Eq. \eqref{eq:sw}) to the training of the baseline FSAF module. Results are reported in Table \ref{tab:ablation} and \ref{tab:eta}. Soft-weighted anchor points offer a significant improvement (up to 1.1\% AP) over the baseline, being insensitive to various hyperparameters. More importantly, AP$_{75}$ is increased by 1.6\%, indicating better localization accuracy at a strict IoU threshold. We also visualize the effect of our soft-weighting scheme in Figure \ref{fig:softweight}(b). The precise box (marked as bold) is kept while the other poorly localized boxes are suppressed, reducing the false attention issue effectively.

\begin{figure}
    \centering
    \includegraphics[width=\columnwidth]{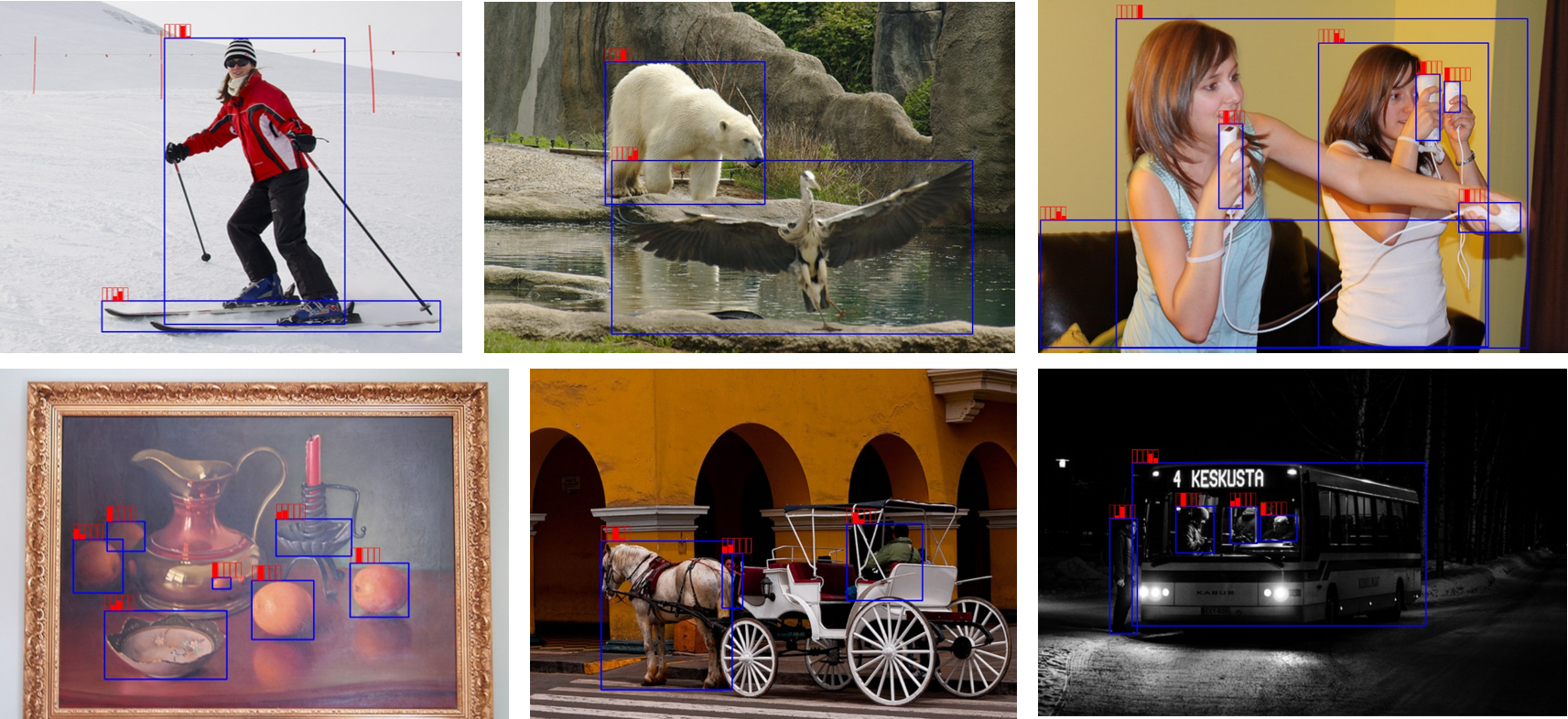}
    \caption{Visualization of the soft feature selection weights from the feature selection network. Weights (the top-left red bars) ranging from 0 to 1 of five pyramid levels ($P_3$ to $P_7$) are predicted for each instance (blue box). The more filled a red bar is, the higher the weight. \textit{Best viewed in color when zoomed in.}}
    \label{fig:vis_ss}
\end{figure}

\textbf{Soft-selected pyramid levels utilize the feature power better.}
Next, we further apply the soft feature selection on top of the soft-weighting scheme, so that each anchor point is down-weighted as in Eq. \eqref{eq:sw+ss}. Table \ref{tab:ablation} and \ref{tab:topk} reports the ablative results. We find that as long as each instance is assigned to more than one pyramid level, we can observe robust $\sim$1.0\% absolute AP improvements over the FSAF module plus soft-weighted anchor points. This indicates that allowing instances to optimize multiple pyramid levels is essential to utilize the feature power as much as possible. Empirically, we assign each instance to the top 3 feature levels with the minimal instance-dependent losses according to Table \ref{tab:topk}. To understand how does the feature selection network assign instances, we visualize the predicted soft selection weights in Figure \ref{fig:vis_ss}. It turns out that larger instances tend to be assigned high weights for higher pyramid levels. The majority of instances can be learned with no more than two levels. Very rare instances need to be modeled by more than two levels, e.g. the sofa in the top right sub-figure of Figure \ref{fig:vis_ss}. This is consistent with the results in Table \ref{tab:topk}.

\textbf{Joint training of the feature selection network has a negligible effect on performance.} As shown in Figure \ref{fig:softselect}, the feature selection network takes in feature extracted from the shared feature pyramid and is jointly trained with the detector. One may argue that the performance improvement by the soft-selected pyramid levels is due to the multi-task learning of the feature selection network and the detector. We prove this is not the case. We conduct an experiment in which the feature selection network is jointly trained with the detector but its predicted soft selection weights are not used. In other words, the weights for anchor points are still following Eq. \eqref{eq:sw} and the feature selection strategy is the same as the baseline FSAF module. It turns out the final AP is 37.1\%, only 0.1\% higher than the 2nd entry and 0.9\% lower than the 3rd entry in Table \ref{tab:ablation}. This means that the major contribution of the soft-selected pyramid levels is actually from \textit{softly selecting multiple levels} rather than the multi-task learning effect from the feature selection network.

\textbf{Our training strategy works well with augmented feature pyramids.}
Different from key-point detectors that use a single high-resolution feature map, the SAPD can enjoy the merits brought by the advanced designs of feature pyramids. Here we adopt the Balanced Feature Pyramid (BFP) \cite{libra-rcnn} and achieve further improvement. The BFP pushes our model with ResNet-50 to a 38.8\% AP, which is 2.9\% higher than the baseline FSAF module. More importantly, our proposed training strategy can robustly work with advanced feature pyramids, offering a steady 2\% AP gain (see 4th and 5th entries in Table \ref{tab:ablation}).

\begin{SCtable}[]
\centering
\setlength\tabcolsep{4pt}
\begin{tabular}{c|c|c c c}
\hline\hline
Backbone & Method & AP & AP$_{50}$ & FPS \\ \hline
\multirow{3}{*}{R-50} & RetinaNet & 35.7 & 54.7 & 11.6 \\ 
 & AB+FSAF & 37.2 & 57.2 & 9.0 \\ 
 & SAPD(Ours) & \textbf{38.8} & \textbf{58.7} & \textbf{14.9} \\ \hline
\multirow{3}{*}{R-101} & RetinaNet & 37.7 & 57.2 &  8.0 \\ 
 & AB+FSAF & 39.3 & 59.2 & 7.1 \\ 
 & SAPD(Ours) & \textbf{41.0} & \textbf{60.7} & \textbf{11.2} \\ \hline
\multirow{3}{*}{X-101-64x4d} & RetinaNet & 39.8 & 59.5 &  4.5 \\  
 & AB+FSAF & 41.6 & 62.4 & 4.2 \\
 & SAPD(Ours) & \textbf{43.1} & \textbf{63.7} & \textbf{6.1} \\ \hline
\end{tabular}
\caption{Head-to-head comparisons of anchor-based RetinaNet, anchor-based plus FSAF module, and our purely anchor-free SAPD with different backbone networks on the COCO \texttt{val2017} set. \textbf{AB}: Anchor-based branches. \textbf{R}: ResNet. \textbf{X}: ResNeXt.}
\label{tab:backbone}
\end{SCtable}

\textbf{SAPD is robust and efficient.}
Our SAPD can consistently provide robust performance using deeper and better backbone networks, while at the same time keeping the detection head as simple as possible. We report the head-to-head comparisons with anchor-based RetinaNet and the more complex anchor-based plus FSAF detector in terms of detection accuracy and speed in Table \ref{tab:backbone}. Except for the head architectures, all other settings are the same. All detectors run on a single GTX 1080Ti GPU with CUDA 10 using a batch size of 1. It turns out that our SAPD gets both sides of two worlds. Our SAPD with purely anchor-free heads can not only run faster than the anchor-based counterparts due to simpler head architecture, but also outperform the \textit{combination} of anchor-based and anchor-free heads by significant margins, i.e. 1.6\%, 1.7\%, and 1.5\% absolute AP increases on ResNet-50, ResNet-101, and ResNeXt-101-64x4d backbones respectively.

\subsection{Comparison to State of the Art}
\label{sec:exp:sota}
We evaluate our complete SAPD on the COCO \texttt{test-dev} set to compare with recent state-of-the-art anchor-free and anchor-based detectors. All of our models are trained using scale jitter by randomly scaling the shorter side of images in the range from 640 to 800 and for 2$\times$ number of epochs as the models in Section \ref{sec:exp:ablation} with the learning rate change points scaled proportionally. Other settings are the same as Section \ref{sec:exp:ablation}.

For a fair comparison, we report the results of single-model single-scale testing for all methods, as well as their corresponding inference speeds in Table \ref{tab:sota}. A visualization of the accuracy-speed trade-off is shown in Figure \ref{fig:ap-ms}. The inference speeds are measured by Frames-per-Second (FPS) on the same machine with a GTX 1080Ti GPU using a batch size of 1 whenever possible. A ``n/a'' indicates the case that the method doesn't provide trained models nor self-timing results from the original paper.

Our proposed SAPD pushes the envelope of accuracy-speed boundary to a new level. We report the results of two series of the backbone models, one without DCN and the other with DCN. Without DCN, our fastest SAPD version based on ResNet-50 can reach a 14.9 FPS while maintaining a 41.7\% AP, outperforming some of the methods \cite{fpn,retinanet,foveabox,fcos,libra-rcnn} using ResNet-101. With DCN, our SAPD forms an upper envelope of state-of-the-art anchor-free detectors and some recent anchor-based detectors. The closest competitor, RPDet \cite{reppoints}, is 1.0\% AP worse and 15ms slower than ours. Compared to key-point anchor-free detectors \cite{centernet,extremenet,cornernet} using Hourglass, our SAPD enjoys significantly faster inference speed (up to 5$\times$ times) and a 2.5\% AP improvement (47.4\% vs. 44.9\%) over the best key-point detector, CenterNet \cite{centernet}.

\section{Conclusion}
This work studied the anchor-point object detection and discovered the key insight lies in the joint optimization of a group of anchor points both within and across the feature pyramid levels. We proposed a novel training strategy addressing two underexplored issues of anchor-point detection approaches, i.e. the false attention issue within each pyramid level and the feature selection issue across all pyramid levels. Applying our training strategy to a simple anchor-point detector leads to a new upper envelope of the speed-accuracy trade-off.

\begin{table}[t]
\scriptsize
\centering
\begin{tabular}{c c c c c c c c c c}
\hline \hline
Method & Backbone & \begin{tabular}[c]{@{}c@{}}Anchor\\free?\end{tabular} & FPS & AP & AP$_{50}$ & AP$_{75}$ & AP$_{S}$ & AP$_{M}$ & AP$_{L}$ \\
\hline
\multicolumn{10}{l}{\textbf{Multi-stage detectors}} \\
Faster R-CNN w/ FPN \cite{fpn} & R-101 & & 9.9 & 36.2 & 59.1 & 39.0 & 18.2 & 39.0 & 48.2 \\
Cascade R-CNN \cite{cascade-rcnn} & R-101 & & 8.0 & 42.8 & 62.1 & 46.3 & 23.7 & 45.5 & 55.2 \\ 
GA-Faster-RCNN \cite{guidedanchor} & R-50 & \checkmark & 9.4 & 39.8 & 59.2 & 43.5 & 21.8 & 42.6 & 50.7 \\
Libra R-CNN \cite{libra-rcnn} & R-101 & & 9.5 & 41.1 & 62.1 & 44.7 & 23.4 & 43.7 & 52.5 \\
Libra R-CNN \cite{libra-rcnn} & X-101-64x4d & & 5.6 & 43.0 & 64.0 & 47.0 & 25.3 & 45.6 & 54.6 \\
RPDet \cite{reppoints} & R-101 & \checkmark & 10.0 & 41.0 & 62.9 & 44.3 & 23.6 & 44.1 & 51.7 \\
RPDet \cite{reppoints} & R-101-DCN & \checkmark & 8.0 & 45.0 & 66.1 & 49.0 & 26.6 & 48.6 & 57.5 \\
TridentNet \cite{tridentnet} & R-101 & & 2.7 & 42.7 & 63.6 & 46.5 & 23.9 & 46.6 & 56.6 \\ 
TridentNet \cite{tridentnet} & R-101-DCN & & 1.3 & 46.8 & 67.6 & 51.5 & 28.0 & 51.2 & 60.5 \\ 
\hline
\multicolumn{10}{l}{\textbf{Single-stage detectors}} \\
RetinaNet \cite{retinanet} & R-101 & & 8.0 & 39.1 & 59.1 & 42.3 & 21.8 & 42.7 & 50.2 \\ 
CornerNet \cite{cornernet} & HG-104 & \checkmark & 3.1 & 40.5 & 56.5 & 43.1 & 19.4 & 42.7 & 53.9 \\
AB+FSAF \cite{fsaf} & R-101 & & 7.1 & 40.9 & 61.5 & 44.0 & 24.0 & 44.2 & 51.3 \\ 
AB+FSAF \cite{fsaf} & X-101-64x4d & & 4.2 & 42.9 & 63.8 & 46.3 & 26.6 & 46.2 & 52.7 \\
GA-RetinaNet \cite{guidedanchor} & R-50 & \checkmark & 10.8 & 37.1 & 56.9 & 40.0 & 20.1 & 40.1 & 48.0 \\
ExtremeNet \cite{extremenet} & HG-104 & \checkmark & 2.8 & 40.2 & 55.5 & 43.2 & 20.4 & 43.2 & 53.1 \\
FoveaBox \cite{foveabox} & X-101 & \checkmark & n/a & 42.1 & 61.9 & 45.2 & 24.9 & 46.8 & 55.6 \\
FCOS \cite{fcos} & R-101 & \checkmark & 9.3 & 41.5 & 60.7 & 45.0 & 24.4 & 44.8 & 51.6 \\ 
FCOS \cite{fcos} w/ imprv & X-101-64x4d & \checkmark & 5.4 & 44.7 & 64.1 & 48.4 & 27.6 & 47.5 & 55.6 \\
CenterNet \cite{centernet} & HG-104 & \checkmark & 3.3 & 44.9 & 62.4 & 48.1 & 25.6 & 47.4 & 57.4 \\
FreeAnchor \cite{freeanchor} & R-101 & & 9.1 & 43.1 & 62.2 & 46.4 & 24.5 & 46.1 & 54.8 \\
FreeAnchor \cite{freeanchor} & X-101-32x8d & & 5.4 & 44.8 & 64.3 & 48.4 & 27.0 & 47.9 & 56.0 \\
SAPD (Ours) & R-50 & \checkmark & 14.9 & 41.7 & 61.9 & 44.6 & 24.1 & 44.6 & 51.6 \\
SAPD (Ours) & R-101 & \checkmark & 11.2 & 43.5 & 63.6 & 46.5 & 24.9 & 46.8 & 54.6 \\
SAPD (Ours) & X-101-64x4d & \checkmark & 6.1 & 45.4 & 65.6 & 48.9 & 27.3 & 48.7 & 56.8 \\
SAPD (Ours) & R-50-DCN & \checkmark & 12.4 & 44.3 & 64.4 & 47.7 & 25.5 & 47.3 & 57.0 \\
SAPD (Ours) & R-101-DCN & \checkmark & 9.1 & 46.0 & 65.9 & 49.6 & 26.3 & 49.2 & 59.6 \\
SAPD (Ours) & X-101-64x4d-DCN & \checkmark & 4.5 & 47.4 & 67.4 & 51.1 & 28.1 & 50.3 & 61.5 \\
\hline
\end{tabular}
\caption{\textit{Single-model and single-scale} accuracy and inference speed of SAPD vs. recent state-of-the-art detectors on the COCO \texttt{test-dev} set. FPS is measured on the same machine with a single \textit{GTX 1080Ti} GPU using the official source code whenever possible. ``n/a'' means that both trained models and timing results from original papers are not available. \textbf{R}: ResNet. \textbf{X}: ResNeXt. \textbf{HG}: Hourglass. For a visualized comparison, please refer to Figure \ref{fig:ap-ms}.}
\label{tab:sota}
\end{table}

\appendix
\section{Discussion}
\label{sec:app}
Besides the ablation studies of the main paper, we ask more questions and conduct additional experiments to further understand our proposed training strategy for SAPD. We follow the same experimental setting as in the ablation studies in Section \ref{sec:exp:ablation}. All models are using the ResNet-50 \cite{resnet} backbone.

\subsection{Soft-Weighting during Training or Testing?}
Previous work like FCOS \cite{fcos} applied soft-weighting in testing. Specifically, FCOS predicts the ``center-ness'' masks from extra network branches and the final score is computed by multiplying the predicted center-ness with the corresponding classification score. Differently, our soft-weighting scheme is applied during the training phase to down-weight anchor points' contribution to the network loss. In other words, FCOS is trained to predict the ``center-ness'' function but we are using the function to directly reweight the loss of anchor points.

\begin{table}[]
    \centering
    \setlength\tabcolsep{2pt}
    \begin{tabular}{c|c c c|c c c c c c}
        \hline\hline
         & SW(ours) & CN(on cls.) & CN(on reg.) & AP & AP$_{50}$ & AP$_{75}$ & AP$_{S}$ & AP$_{M}$ & AP$_{L}$ \\ \hline
         FSAF \cite{fsaf} &  &  & & 35.9 & 55.0 & 37.9 & 19.8 & 39.6 & 48.2 \\
         & \checkmark &  &  & \textbf{37.0} & 55.8 & 39.5 & 20.5 & 40.1 & 48.5 \\ 
         &  & \checkmark & & 36.1 & 55.2 & 38.1 & 20.3 & 40.0 & 47.4 \\
         &  &  & \checkmark & 36.5 & 55.5 & 38.9 & 21.0 & 40.1 & 48.3 \\
         & \checkmark &  & \checkmark & 36.8 & 55.2 & 39.4 & 20.6 & 40.2 & 48.0 \\
         \hline
    \end{tabular}
    \caption{Performance comparison between soft-weighting the loss during training by soft-weighted anchor points and down-weighting the confidence score during testing by predicted ``center-ness''. \textbf{SW}: soft-weighted anchor points, \textbf{CN}: center-ness, ``on cls.'': center-ness branch on the classification branch, ``on reg.'': center-ness branch on the regression branch.}
    \label{tab:centerness}
\end{table}

For comparison between soft-weighting in training vs. in testing, we implement the ``center-ness'' mask branches attached to our baseline FSAF module \cite{fsaf} using the official code and optimize them the same way as \cite{fcos}. Performances are reported in Table \ref{tab:centerness}. Our soft-weighting scheme in training is more effective than the various versions of center-ness weighting in testing. The best version of center-ness improves the AP by 0.6\% while our soft-weighting scheme achieves a 1.1\% AP gain. We think the reason is that soft-weighting during training is directly addressing the false attention issue, in which anchor points with poorly aligned features for precise localization are down-weighted. But in center-ness weighting, the anchor points are still contributing equally to the network loss, which is forcing all anchor points to perform equally well no matter how good are their feature representations. So the soft-weighting during testing is not fully resolving the false attention issue. This is further verified by the fact that if our soft-weighting scheme is applied on top of the center-ness weighting we can observe another improvement (see 4th and 5th entries in Table \ref{tab:centerness}). However, if we compare between 2nd and 5th entries, applying center-ness weighting on top of our soft-weighting scheme is not improving the performance, which indicates that our soft-weighting scheme alone can work well to suppress the poorly localized detections.
Therefore, we believe reweighting the anchor loss is more close to the essence of suppressing poorly localized detections than reshaping the confidence score during inference.

\subsection{Which Loss to Reweight?}
In Eq. (2) of the main paper, the anchor point loss is the summation of the classification focal loss and the localization IoU loss for positive samples. By default, our soft-weighting scheme is applied to both classification and localization losses in the soft-weighted anchor points. In this section, we study the effect of applying our soft-weighting scheme to only the classification (cls) loss or the localization (loc) loss. Results are reported in Table \ref{tab:reweight}. If we only reweight the single cls loss or loc loss, the performance becomes even worse than the baseline. The possible reason is that down-weighting a loss causes the network to focus on optimizing the other unweighted loss and the network is biased to be good at a single task. But the detection problem requires the network to be balanced for both proper classification and localization abilities. 

\begin{table}[]
\centering
\setlength\tabcolsep{2pt}
\begin{tabular}{c|cc|cccccc}
\hline \hline
        & cls        & loc        & AP & AP$_{50}$ & AP$_{75}$ & AP$_S$ & AP$_M$ & AP$_L$ \\ \hline
FSAF \cite{fsaf}    &            &            & 35.9 & 55.0 & 37.9 & 19.8 & 39.6 & 48.2 \\
FSAF+SW & \checkmark &            & 33.3 & 50.6 & 35.4 & 18.3 & 37.7 & 43.1 \\
FSAF+SW &            & \checkmark & 35.6 & 55.2 & 37.6 & 19.7 & 39.7 & 45.8 \\
FSAF+SW & \checkmark & \checkmark & \textbf{37.0} & 55.8 & 39.5 & 20.5 & 40.1 & 48.5 \\ \hline
\end{tabular}
\caption{The effect of applying our soft-weighting scheme to only the classification (cls) loss, or only the localization (loc) loss, or the summation of classification and regression loss (cls+loc) for the soft-weighted anchor points. \textbf{SW}: soft-weighed anchor points.}
\label{tab:reweight}
\end{table}

Therefore, compared to previous detection methods that reweighting only the classification loss \cite{retinanet,cornernet,cl}, our soft-weighting scheme is more comprehensive and balanced. 

\section{Visualization of Feature Selection Network}
\begin{figure}
    \centering
    \includegraphics[width=\columnwidth]{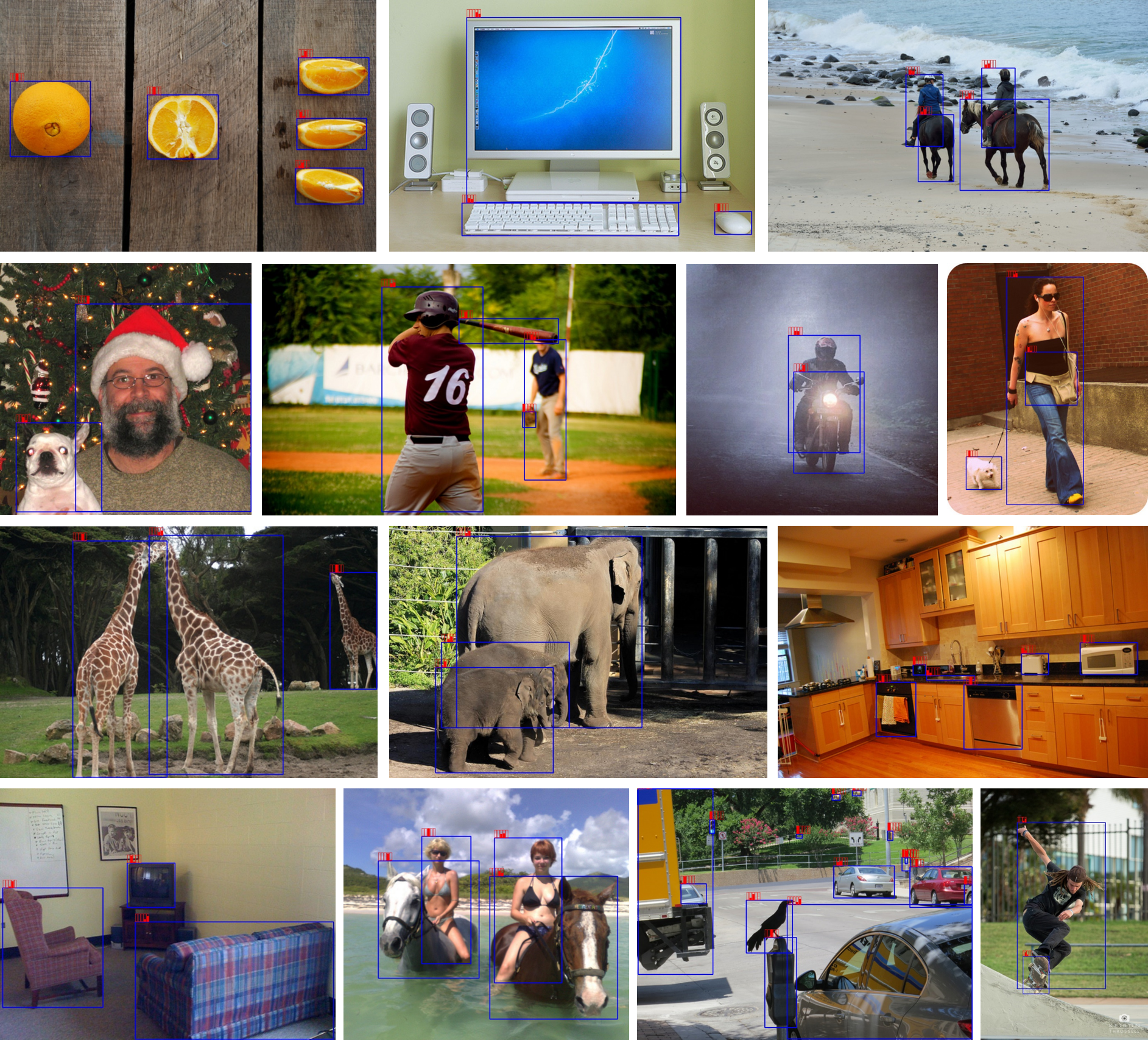}
    \caption{More visualization of the soft-selection weights from the feature selection network. Weights (the top-left red bars) ranging from 0 to 1 of five pyramid levels ($P_3$ to $P_7$) are predicted for each instance (blue box). The more filled a red bar is, the higher the weight is. \textit{Best viewed in digital version and zoomed in.}}
    \label{fig:fsn}
\end{figure}

We visualize more examples of the soft-selected pyramid levels from the feature selection network in Figure \ref{fig:fsn}. The feature selection network predicts the per-level ``participation'' degree for each instance to fully explore the power of feature pyramid and it is agnostic to the instance class, being general for a variety of objects including animals, human, food, vehicle, furniture, etc. Using features from multiple pyramid levels for detection is better than the online feature selection strategy in the FSAF module \cite{fsaf}, which only chooses a single level to assign the instance when training the network. 

\clearpage
%
%
\bibliographystyle{splncs04}
\bibliography{egbib}
\end{document}